\documentclass[conference]{IEEEtran}
\IEEEoverridecommandlockouts
\usepackage{amsmath,amssymb,amsfonts}
\usepackage{algorithmic}
\usepackage{graphicx}
\usepackage{textcomp}
\usepackage{xcolor}
\usepackage[numbers]{natbib}
\usepackage{stackengine}
\usepackage{multirow}
\usepackage{academicons}
\usepackage{microtype}
\usepackage{natbib}
\definecolor{orcidlogocol}{HTML}{A6CE39}

\def\BibTeX{{\rm B\kern-.05em{\sc i\kern-.025em b}\kern-.08em
    T\kern-.1667em\lower.7ex\hbox{E}\kern-.125emX}}

\begin{document}

\title{TsFeX: Contact Tracing Model using Time Series Feature Extraction and Gradient Boosting\\

}

\author{
\IEEEauthorblockN{Valerio  Antonini}
\IEEEauthorblockA{\textit{School of Computing} \\
\textit{Dublin City University, Ireland}\\
valerio.antonini3@.mail.dcu.ie}
\and

\IEEEauthorblockN{Yingjie Niu}
\IEEEauthorblockA{\textit{School of Computer Science} \\
\textit{University College Dublin, Ireland}\\
yingjie.niu@ucdconnect.ie}
\and

\IEEEauthorblockN{Manuela Nayantara Jeyaraj}
\IEEEauthorblockA{\textit{School of Computing} \\
\textit{Technological University Dublin, Ireland}\\
manuela.n.jeyaraj@mytudublin.ie}
\and

\IEEEauthorblockN{Sonal Santosh Baberwal}
\IEEEauthorblockA{\textit{School of Electronic Engineering} \\
\textit{Dublin City University, Ireland}\\
sonal.baberwal2@mail.dcu.ie}
\and

\IEEEauthorblockN{Faithful Chiagoziem Onwuegbuche}
\IEEEauthorblockA{\textit{School of Computer Science} \\
\textit{University College Dublin, Ireland}\\
faithful.chiagoziemonwuegb@ucdconnect.ie}
\and

\IEEEauthorblockN{Robert Foskin}
\IEEEauthorblockA{\textit{School of Computer Science} \\
\textit{University College Dublin, Ireland}\\
robert.foskin@ucdconnect.ie}
}

\maketitle

\begin{abstract}
With the outbreak of COVID-19 pandemic, a dire need to effectively identify the individuals who may have come in close-contact to others who have been infected with COVID-19 has risen. This process of identifying individuals, also termed as ‘Contact tracing’, has significant implications for the containment and control of the spread of this virus. However, manual tracing has proven to be ineffective calling for automated contact tracing approaches. As such, this research presents an automated machine learning system for identifying individuals who may have come in contact with others infected with COVID-19 using sensor data transmitted through handheld devices. This paper describes the different approaches followed in arriving at an optimal solution model that effectually predicts whether a person has been in close proximity to an infected individual using a gradient boosting algorithm and time series feature extraction. 
\end{abstract}


\section{Introduction}

The COVID-19 pandemic has been a challenge and a shift on personal and global levels in terms of health (physical and mental well-being), economy, lifestyle and society \cite{ferretti2020quantifying}. To fight this battle and overcome the infection by minimizing the contact with infected individuals, contact tracing can be implemented \cite{he20202}. The goal of contact tracing is to detect and alert those who would have come in close proximity to a probable case of an infectious disease or an infected person as a means of controlling the spread of infection. The tracing can be done manually or automatically with the help of mobile phones and inbuilt sensors. However, the severity of the COVID-19 pandemic and the high proportion of transmissions means manual contact tracing may not be effective in controlling the spread of the disease hence, the need for automatic contact tracing. 

Existing literature works have conceptualized using Bluetooth sensors (Bluetooth Low Energy signals) with a goal of detecting whether the individual is too close to any infected person. However, the RSSI (Received Signal Strength Indicator) value is noisy compared to the actual distance in real environments \cite{shankar2020proximity}. Motivated by the work of \cite{met_ref_1}, this research uses sensor data such as accelerometer, altitude and gyroscope in addition to the Bluetooth data to evaluate the plausible application of Machine Learning algorithms to compute the distance between sensors, that is, mobile devices, and determine if an individual has been in close contact to an infected individual.  

\section{Related Work}

Sensors embedded in the smartphone provide the raw data necessary to estimate whether an event can be defined as 'contact' or 'no contact'. \cite{he20202} proposes a two-stage model in which raw data is transformed into a fixed-length vector in stage one and  subsequently, a pre-trained deep learning classifier is used to predict the type of event such as an event of 'contact 'no-contact' with the virus by means of the fixed-length vector in stage two..
\par 
Similarly, \cite{shankar2020proximity} focuses on the temporal characteristics of the dataset. The task was modelled as a time series approach and the data was broken down into 150 time-steps for each 4-sec chunk and later trained using several deep learning models, Decision Tree-based models, Support Vector Machines (SVM) and Naive Bayes Classifier. Their results show that a temporal, one-dimensional convolutional network achieves the best performance though the SVM and tree-based models were training on a smaller subset of the data.
\par 
\cite{met_ref_1} proposed a few different possible approaches out of which the authors had implemented the approach that involved condensing the information in the event files using properties recorded by the transmitter and receiver ends along with the RSSI values. According to the authors, this task's main challenge was in standardizing all the events and establishing features that can be efficiently utilised in a machine learning approach. The results showed that the Gradient Boosting Machine (GBM) and Multi-Layer Perceptron (MLP) algorithms trained with handcrafted features were effective in estimating the distance between two phones. It further emphasised the need for incorporating domain knowledge in the training of machine learning models. However, the work of \cite{met_ref_1} presented the prospect to utilise those supplementary phone sensor data in the estimation of the distance between two phones. 

\section{Task Description}
The National Institute of Standards and Technology (NIST) has launched a project TC4TL (Too Close for Too Long) to explore promising ideas of using Bluetooth and other sensor signals to evaluate the performance of the state-of-the-art models to address this issue of contact tracing.
\par
Accordingly, notifying individuals about possible virus exposure around them in order to limit the infection is one of the key aspects in containing the pandemic as quoted by NIST \cite{nisteval}; “Current approaches to automated exposure notification rely on using Bluetooth Low Energy (BLE) signals (or chirps) from smartphones to detect if a person has been too close for too long (TC4TL) to an infected individual.”
\par
Apart from the RSSI values obtained being too noisy, other factors such as the type of environment and the body and phone positions can affect the results. The goal is to effectively and accurately measure the distance between the two devices in a real world environment. The goal of this challenge is to estimate the time and distance between two cellphones using sensor data such as BLE signals, accelerometer, gyroscope, etc., as input and evaluating the output to be a TC4TL or non-TC4TL event (NIST TC4TL challenge).
\par
With respect to the previously established state-of-the-art model \cite{met_ref_1}, the limitation they faced was in effectively harnessing the inertial measurement units (IMU) data for the distance predictions. Hence, this paper addresses this open problem stated as a limitation and future direction in \cite{met_ref_1}. 
\par 
Therefore, the main contribution of this paper is the proposal of a contact tracing model that leverages IMU sensor data using time series feature extraction and gradient boosting algorithm tuned using Bayesian optimization. This model has demonstrated comparable performance on par with the current state-of-the-art model for the same task and outperforms the state-of-the-art on the coarse grain dataset.

\section{Methodology}

\subsection{Data exploration}

Motivated by the work of \cite{met_ref_1}, the NIST data was explored. This data was collected between two sensor enabled devices. It comprises of 15,552 training files, 936 development files and 8,423 test files out of which bluetooth, accelerometer, altitude and gyroscope are provided in addition to the information about the devices, their locations and activities, which were used for training and thereafter, testing. These features are described in Table \ref{tab:tab1}

\begin{table}[!ht]
\resizebox{0.48\textwidth}{!}{%
\begin{tabular}{lll}
Input &
  Data &
  Description \\ \hline
Predicted distance &
  \begin{tabular}[c]{@{}l@{}}Continuous\\ (1,2,4,5)\end{tabular} &
  \begin{tabular}[c]{@{}l@{}}Distance estimate. Coarse grain\\ {[}TX:-52, N:2.6{]}, Fine grain: \\ {[}TX:-54, N:2.1{]}. RSSI is the \\ mean RSSI signal strength for the file.\end{tabular} \\ \hline
\begin{tabular}[c]{@{}l@{}}Normalized* \\ mean RSSI\end{tabular} &
  Continuous (0,1) &
  Mean RSSI signal strength for the file \\ \hline
\begin{tabular}[c]{@{}l@{}}Normalized* \\ path loss attenuation\end{tabular} &
  Continuous (0,1) &
  Transmit power - 41 - mean RSSI \\ \hline
Coarse/fine grain &
  Categorical (0,1) &
  0 for fine graine and 1 for coarse grain \\ \hline
TXPower &
  Categorical (0,2) &
  \begin{tabular}[c]{@{}l@{}}0 for TXPower 7 / unknown,\\ 1 indicates TXPower 8 and 2\\ indicates TXPower 12\end{tabular} \\ \hline
TXCarry &
  Categorical (0,2) &
  \begin{tabular}[c]{@{}l@{}}0 for unknown transmitter carry state,\\ 1 for hand and 2 for pocket\end{tabular} \\ \hline
RXCarry &
  Categorical (0,2) &
  \begin{tabular}[c]{@{}l@{}}Same encoding as TXCarry for receiver\\ phones\end{tabular} \\ \hline
TXPose &
  Categorical (0,2) &
  \begin{tabular}[c]{@{}l@{}}0 for unknown transmitter pose,\\ 1 for sitting and 2 for standing\end{tabular} \\ \hline
RXPose &
  Categorical (0,2) &
  \begin{tabular}[c]{@{}l@{}}Same encoding as TXPose\\ but for receiver phones\end{tabular} \\ \hline
TXDevice &
  Categorical (0,2) &
  \begin{tabular}[c]{@{}l@{}}Model of the transmitter phone. 0 for \\ unknown/ older than iPhone 7, 1 for \\ iPhone 7-8, 2 for iPhone X or newer\end{tabular} \\ \hline
RXDevice &
  Categorical (0,2) &
  \begin{tabular}[c]{@{}l@{}}Model of the receiver phone. Same \\ encoding as TXDevice\end{tabular}
\end{tabular}%
}
\vspace{3mm}
\caption{Baseline features}
\vspace{-5mm}
\label{tab:tab1}
\end{table}

Each train and test file is considered to be a 'contact event' which itself is comprised of a number of individual measurement events or 'looks' corresponding to 4 second periods of continuous data recording. The distribution of the number of individual looks per data file is shown in Figure \ref{fig:fig1} below.

\vspace{-2mm}
\begin{figure}[!ht]
\centering
\includegraphics[width=0.5\textwidth]{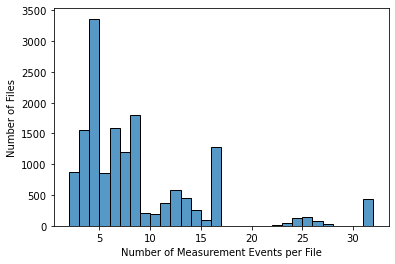}
\caption{Distribution of Measurement Intervals from the Contact Event Files.}
\vspace{2mm}
\label{fig:fig1}
\end{figure}

\vspace{-3mm}
\subsection{Models and Approaches}
Due to the wide array of solution models that have already been proposed for this task of contact tracing, initially, a baseline model was reproduced according to the approach outlined by \cite{met_ref_1} representing the actual best result according to the post-challenge leaderboard on the official TC4TL Challenge site. Among all the sensor data available, only the RSSI signal data has been used in addition to some information regarding the devices. The sensor data and training instances were processed with a script that standardised the data, normalised the variables and retained useful features to be used within the machine learning model (the list of the features composing the final dataset is shown in Table \ref{tab:tab1}). Having identified the variables, two \textit{Gradient Boosting Machine} classifiers were implemented; one for the fine grain and another for the coarse grain data.
\vspace{5mm}

\subsubsection{Approaches}
\paragraph{Baseline} Having reproduced the state-of-the-art and developing on the idea of using gradient boosting algorithms, in this approach the same baseline features were implemented using an XGBoost classifier instead of a gradient boosting machine (GBM) or multi-layer perceptron (MLP) as opposed to the baseline. Though XGBoost and GBMs both adhere to the principle of gradient boosting, xgboost uses a more regularized formalization technique that prevents the model from over-fitting. In addition, though the entire ensemble cannot be parallelized as each tree is dependent on the preceding tree, the process of building the nodes within each tree can be parallelized making xgboost faster than GBM. Hence, the initial step was to observe the xgboost classifier’s performance on the same set of features as the baseline model.

\paragraph{Approach 1} With access to the IMU sensor data that includes the accelerometer and gyroscope readings as time series data, the K-Shape algorithm  was used to introduce unsupervised time series clustering as the next strategy. on the sensor data which generates homogeneous and thoroughly segregated clusters using a scalable and iterative optimization technique \cite{met_ref_2}. Unlike conventional clustering algorithms, K-Shape is not affected by shift or scale changes on the input time-series sequences. It does not require domain specific information or knowledge of the underlying physics of the data. Instead, it finds clusters without regard to the sequencing order, time scale factors, or shift parameters, while it takes into account signal correlations among sequences. Hence, K-Shape was considered as the most viable time series clustering algorithm given the limited background information on the sensor data. In doing so, the multivariate Accelerometer (acc) data had to be transformed into as a univariate single channel of data.
\par 
In this approach, padding was used to address the size difference of the time series data.
\par 
The cluster outputs were evaluated using the inertia which measures the internal coherence between the clusters using the following formula.
\vspace{-2mm}
\begin{align}
    inertia = \sum_{i=0}^{n}\min_{\mu_j \in C}(\| x_i -\mu_j \|^2)
\end{align}

\vspace{2mm}

Here, $n$ represents the samples, $C$ indicates the center of each cluster which is interpreted by the average cluster centroid, $mu_j$. Hence, the inertia measures the sum of the squared difference of the distance of each sample in a cluster to its centroid.
\par 
Though the inertia is not a normalized metric, the choice of this measure for this cluster evaluation was based on the fact that, as opposed to the inertia, other metrics such as the Rand index, Fowlkes-Mallows index and Mutual information based scores all require the knowledge of the ground truth classes.
\par
As such, the number of clusters that were found to be optimal was 14. 
\par 
Hence, in this approach, in addition to the baseline features, the cluster label was provided as a feature and trained using the XGBoost model. However, this approach yielded a higher nDCF. The use of zero padding to address the length difference in the time series data could possibly be introducing noise through such an interpolation resulting in a higher nDCF value for the fine and coarse grained data.

\paragraph{Approach 2} In order to resolve the issue in approach 1, the model was updated to use a re-sampling technique instead of padding to smooth out the length difference in the time series data. Re-sampling converts the frequency and re-samples the time series data. The frequency of the data indicates the time where each value in the series is associated with a number representing how many seconds ago it occurred. Hence, re-sampling is sought when data is coming in from multiple times, measurements are repeated at set intervals and/or while performing statistical analysis on the data. As such, the \texttt{TimeSeriesResampler} made available within the \textit{tslearn} python package was used. The gap across time series of different varying lengths was computed via dynamic time warping.
\par 
With the K-Means clustering done by re-sampling the data, the best performance was achieved with 4 clusters whereas, increasing the number of clusters showed detrimental performance scores.
\par 
Following up on the 2\textsuperscript{nd} approach, more IMU data such as the accelerometer, gyroscope, attitude, gravity and magnetic field were introduced into the model. This was done by generating 5 time series clusters for the respective IMU features and passing those cluster labels as features into the model. 

\paragraph{Approach 3} The next approach opted to explore the use of time series features. As such, features were drawn from this time series data using Rocket transformer \cite{met_ref_6}. ROCKET is an abbreviation for RandOm Convolutional KErnel Transform. ROCKET randomly produces many convolutional kernels and derives two characteristics from them: the maximum and the fraction of positive values. The advantage of Rocket comes from its usage of convolutional kernels. Unlike other methods used for classification on time series which consider only one representative type, for example, the shape, the variance of the signals or the frequency, Rocket can simultaneously represent different types of multiple characteristics. Moreover, in terms of computational time and performance, Rocket is significantly fast and accurate. 
\par 
The model took in all the IMU features without pre-processing and re-sampling was used to smooth the length differences in the time series data. The features extracted by Rocket were processed using the Ridge classifier which is a linear model and by default it performs the leave-one-out cross validation. The ridge classifier initially transforms binary targets and proceeds on as a regression task.
\par 
In addition to the Ridge classifier, the XGBoost model has been applied on top of the set of features extracted by ROCKET.

\paragraph{Approach 4}
This approach follows the idea, suggested in \cite{met_ref_4} of employing two distinct models for coarse grain and fine grain data. Furthermore, two different sets of features are used for the subgroups. The hypothesis is that some features could likely be more useful for making predictions on one sub-group rather than the other. Here, the features selection has been carried out by assessing the predictions obtained with various combinations of features through the evaluation software. As such, the features chosen for the fine grain and coarse grain data are described here below.
\begin{itemize}
  \item \textit{\textbf{Fine-grain data}}:
\end{itemize}
Time series data has recently emerged as one of the most challenging to process in machine learning due to its high computational cost and massive datasets. Thus, we resolved to using tsfresh to automatically extract numerous features, describing specific aspects about the time series \cite{met_ref_5}. Examples of extracted features are:
\begin{itemize}
  \item \textit{count\_above(x,s)}: proportion of values in x higher than s.
  \item \textit{number\_cwt\_peaks(x)}: number of peaks in x
  \item \textit{percentage\_of\_reoccurring\_values\_to\_all\_values(x)}: proportion of entries in the time series that appear more than once
\end{itemize}
A sample of the features harnessed from the time series is shown in Table \ref{tab:tab2}. In addition, the entire list of features that are extracted by tsfresh is provided in the tsfresh documentation\footnote{ https://tsfresh.readthedocs.io/en/latest/index.html}.
\par 
The dataset for fine grain data is composed of the baseline features plus the features extracted by tsfresh from the following sensor data: RSSI, accelerometer, gyroscope and attitude. Since the last three sensors collect the data over dimensional coordinates (x,y,z), each coordinate has been treated as a single time series and the features have been extracted for each coordinate. Those features turned out having only one value removed.

\begin{table}[!ht]
\centering
\resizebox{0.48\textwidth}{!}{%
\begin{tabular}{ll}
\textbf{Feature} &
  \textbf{Description} \\ \hline
\multicolumn{1}{l|}{Energy} &
  \begin{tabular}[c]{@{}l@{}}Absolute energy of the time series: sum over\\ the squared values\end{tabular} \\ \hline
\multicolumn{1}{l|}{\begin{tabular}[c]{@{}l@{}}Absolute \\ Maximum\end{tabular}} &
  Highest absolute value of the time series \\ \hline
\multicolumn{1}{l|}{\begin{tabular}[c]{@{}l@{}}Count Above\\ Mean\end{tabular}} &
  \begin{tabular}[c]{@{}l@{}}The number of values of the time series higher\\ than the average of the time series\end{tabular} \\ \hline
\multicolumn{1}{l|}{\begin{tabular}[c]{@{}l@{}}Fourier\\ Entropy\end{tabular}} &
  \begin{tabular}[c]{@{}l@{}}Calculate the binned entropy of the power spectral\\ density of the time series\end{tabular} \\ \hline
\multicolumn{1}{l|}{Kurtosis} &
  \begin{tabular}[c]{@{}l@{}}Kurtosis of the time series calculated with the adjusted\\ Fisher-Pearson standardized moment coefficient G2\end{tabular} \\ \hline
\multicolumn{1}{l|}{\begin{tabular}[c]{@{}l@{}}Longest \\ Strike above\\ the Mean\end{tabular}} &
  \begin{tabular}[c]{@{}l@{}}Length of the longest consecutive subsequence in x \\ that is bigger than the mean of x\end{tabular} \\ \hline
\multicolumn{1}{l|}{\begin{tabular}[c]{@{}l@{}}Variation\\ Coefficent\end{tabular}} &
  \begin{tabular}[c]{@{}l@{}}Variation coefficient: standard error/ mean. Returns the \\ relative value of variation around the mean of the time series\end{tabular}
\end{tabular}%
}

\vspace{3mm}
\caption{A sample of features extracted by tsfresh}
\label{tab:tab2}
\end{table}

\vspace{-5mm}
\begin{itemize}
  \item \textit{\textbf{Coarse-grain data}}:
\end{itemize}
The approach follows the idea of manipulating sensor data in order to obtain initially hidden variables as much as possible that could be potentially useful for the final prediction (such as the feature, \textit{predicted distance} \cite{met_ref_4}, derived from the RSSI and TXPower). Therefore, this feature engineering was applied to the IMU data and specifically, to the accelerometer (ACC), altitude (ALT) and gyroscope (GYR) features. Subsequently, the multivariate time series of ACC and ALT data were transformed using the following equations.
\begin{equation}
    \begin{aligned}
    \label{eqn:Normalized_imu}
        MagnACC = \sqrt{x^2 + y^2 + z^2}\\
        MagnALT = \sqrt{x^2 + y^2}\\
    \end{aligned}
\end{equation}
With regard to the GYR data, three new features were designed: the normalized mean of the values along the x, y and z coordinates, respectively. Table \ref{tab:tab3} shows the normalized and retained IMU features added into the model.

\begin{table}[!ht]
\resizebox{0.48\textwidth}{!}{%
\begin{tabular}{lll}
Feature                                                          & Datatype         & Description                                                                     \\ \hline
\begin{tabular}[c]{@{}l@{}}Normalized \\ mean X GYR\end{tabular} & Continuous (0,1) & \begin{tabular}[c]{@{}l@{}}Mean X coordinate of\\ Gyroscope signal\end{tabular} \\ \hline
\begin{tabular}[c]{@{}l@{}}Normalized \\ mean Y GYR\end{tabular} & Continuous (0,1) & \begin{tabular}[c]{@{}l@{}}Mean Y coordinate of\\ Gyroscope signal\end{tabular} \\ \hline
\begin{tabular}[c]{@{}l@{}}Normalized \\ mean Z GYR\end{tabular} & Continuous (0,1) & \begin{tabular}[c]{@{}l@{}}Mean Z coordinate of\\ Gyroscope signal\end{tabular} \\ \hline
\begin{tabular}[c]{@{}l@{}}Magnitude\\ Acceleration\end{tabular} &
  Continuous (0,N) &
  \begin{tabular}[c]{@{}l@{}}Sqrt of the sum of each\\ squared value of accelerometer\\ over the coordinates (x,y,z)\end{tabular} \\ \hline
\begin{tabular}[c]{@{}l@{}}Magnitude\\ Altitude\end{tabular} &
  Continuous (0,N) &
  \begin{tabular}[c]{@{}l@{}}Sqrt of the sum of each squared\\ value of altitude over the \\ coordinates (x,y)\end{tabular}
\end{tabular}%
}
\vspace{3mm}
\caption{Feature engineering for coarse-grain data}
\vspace{-3mm}
\label{tab:tab3}
\end{table}

These set of features in addition to the baseline features make up the coarse grain dataset.
\par 
Thereafter, the coarse grain dataset and the fine grain dataset are fed into two distinct XGBoost models.
\par 
Both XGBoost models underwent a hyper-parameter tuning using Bayesian optimization \cite{met_ref_6}. Bayesian optimization takes the Bayes theorem \cite{met_ref_7} to identify the optimal maximum or minimum objective function. The search is constrained to the space of possible values for the hyper-parameters. The method is called Bayesian because it works on a probability distribution over the space of possible values, and the optimization process finds a point with inbuilt methods like gradient descent and stochastic methods.
\par 
The problem with most of the other optimization approaches is that they can find local optima and fail to find global optima. Bayesian Optimization overcomes this problem by using Bayes Theorem to model uncertainty about parameter values, even though there is no way to be certain about any particular value in advance. This allows the model to make predictions about where it should look for an optimum next based on previous information learned from exploring different ways of tuning parameters that it already knows.

\subsection{Model Evaluation}

The base goal of the evaluation is to collect a set of contact events composed of various measures such as the RSSI chirps, the location where the sensor was placed during the time of measurement, etc. These events will be used to determine the distance between two phones.
\par 
For each such receiver-transmitter (RX-TX) configuration, the recorded values are analyzed using the phone sensor and RSSI log. Consequently, four-second long windows of varying lengths are used to retrieve the test samples.
\par 
Afterwards, the test samples are split into two batches, namely fine-grain (1.2m, 1.8m, 3m, 4.5m) and coarse-grain (1.8m, 4.5m) distance sets. For example, 1.2m in a fine-grain data instance that represents that the distance between two phones can vary by up to 1.2m with respect to a single contact event. The fine-grain subgroup will use the variable values to define events and compute the performance according to the values 1.2m, 1.8m and 3.0m. In the coarse-grain subgroup, the performance will only be computed with a single distance, that is, 1.8 metres.
\par 
The system predicts the distance between a contact event and the object's intended distance. And based on the predicted distance, the system is able to decide whether a given event is a TC4TL event. Finally, for each contact event, the probability of the misses ($P_{miss}$) and false alarm ($P_{false}$) predictions are calculated as shown in equation \ref{eq:missfa}.

\begin{equation}
    \begin{aligned}
    \label{eq:missfa}
        P_{miss} = \frac{Number\;of\;not\_TC4TL\;events}{Number\;of\;TC4TL\;events}\\
        \\
        P_{false} = \frac{Number\;of\;TC4TL\;events}{Number\;of\;not\_TC4TL\;events}
    \end{aligned}
\end{equation}

\par 
The overall performance of the contact tracing model is determined using the normalized decision cost function (nDCF) as shown in equation \ref{eq:ndcf} derived from the probability of the misses and the false alarms as per equation \ref{eq:missfa}.
\begin{equation}
    \begin{aligned}
    \label{eq:ndcf}
        nDCF = \frac{w_{miss}P_{miss} + w_{false}P_{false}}{min(w_{miss}w_{false})}
    \end{aligned}
\end{equation}

In such a decision making process with multiple possible outcomes, each decision will typically have its own cost function that is indirectly proportional to how desirable each outcome is. The cost function is different from the utility function as it may associate a cost to an outcome which is preferred to another outcome as opposed to just mapping onto a utility value. Hence, the lower the normalized decision cost function value, the better. As such, the model's nDCF is computed using the above method made available by NIST\footnote{NIST TC4TL Challenge: https://tc4tlchallenge.nist.gov/} as a script.
In order to encourage researchers to approach the challenge of contact tracing, NIST maintains a post-challenge leaderboard on the official TC4TL Challenge website, showing the performances of the various teams according to the evaluation score discussed above. The teams are ranked based on their score on the coarse grain data subset.

\section{Results and Discussions}

The results observed through the evaluations conducted in accordance to the description in the preceding section are tabulated here below.

\begin{table}[!ht]
\centering
\resizebox{0.48\textwidth}{!}{%
\begin{tabular}{lll}
\textbf{Approach}           & \textbf{Description}                                                                                                                      & \textbf{nDCF} \\ \hline 
State-of-the-art   & \begin{tabular}[c]{@{}l@{}}Gradient Boosting Machine/\\ Multilayer Perceptron-based\\ model\end{tabular}                         & 0.52 \\ \hline
Baseline                  & \begin{tabular}[c]{@{}l@{}}Baseline features with\\ XGBoost classifier\end{tabular}                                              & 0.60 \\ \hline
1                  & \begin{tabular}[c]{@{}l@{}}Padding time-series data and \\ k-shape clustering on IMU\\ sensor data.\end{tabular}                 & 0.71 \\ \hline
\multirow{1}{*}{2} & \begin{tabular}[c]{@{}l@{}}Resampling timeseries data and\\ k-means clustering\end{tabular}                                      & 0.63 \\ \cline{2-3} 
                   & \begin{tabular}[c]{@{}l@{}}Inclusion of IMU data, resampling\\ time series data, generating time \\ series clusters\end{tabular} & 0.66 \\ \hline
\multirow{2}{*}{3} &
  \begin{tabular}[c]{@{}l@{}}Time series feature extraction using\\ ROCKET\footnote{ROCKET: https://arxiv.org/abs/1910.13051} and use of the ridge classifier\end{tabular} &
  0.94 \\ \cline{2-3} 
                   & \begin{tabular}[c]{@{}l@{}}Time series feature extraction using\\ ROCKET and XGBoost classifier\end{tabular}              & 0.63 \\\hline
4 &
  \begin{tabular}[c]{@{}l@{}}Time series feature extraction using\\ tsfresh\footnote{tsfresh: https://tsfresh.readthedocs.io/en/latest/} for fg data, feature engineering\\for cg data and XGBoost classifier with\\ hyperparameter optimization using\\ Bayesian optimization over the two dfs\end{tabular} &
  \textbf{0.54}
\end{tabular}%
}
\vspace{3mm}
\caption{Model Performance as demonstrated by the different approaches implemented.}
\vspace{-5mm}
\label{tab:tab4}
\end{table}

The transition from one model to another model's implementation approach was based on the intuitive observations, computational costs and performance scores attained by the former approach. 
{In table \ref{tab:tab4}}, the state-of-the-art represents the model developed by \cite{met_ref_1}. As described in their paper, the mean nDCF attained was 0.52. Developing on the idea of using gradient boosting as the baseline, the XGBoost classifier was implemented in place of the gradient boosting machine. However, the mean nDCF rose to 0.60 with this baseline. This unforeseen performance of the XGBoost classifier was attributed to the discrepancies in the data pre-processing methods carried out between that of the state-of-the-art model's original implementation and the baseline approach's pre-processing steps. 
\par 
Given the access to the IMU sensor data, approach 1 leveraged those time series data and shifted to a clustering algorithm through which it achieved a mean nDCF of 0.71. The performance degradation in this approach was due to the padding of time-series data.
\par
Therefore, in approach 2, re-sampling was employed as an alternative to smoothing the length difference in the time-series data while following the clustering approach. Though this seemed to effectively bring the nDCF down to 0.63, a variation of this method was attempted by including the IMU data and generating time series clusters. While the inclusion of the IMU data effectively reduced the nDCF on the coarse grain data, it increased the overall mean nDCF to 0.66. 
\par 
Having gained an intuitive understanding about the role of the time series data in the predictions from approach 2, a feature extraction was done on top of the time series data using ROCKET while reverting to treating the predictions as a classification task in the 3\textsuperscript{rd} approach. However, the use of the ridge classifier worsened the performance with a mean nDCF of 0.94 as it was shrinking the coefficients to 0 and trading the variance for bias. 
\par 
Since the choice of classifier was found to cause the undesirable performance, the same approach was re-implemented with the XGBoost classifier and yielded a mean nDCF of 0.63. 
\par 
The approach reaching the best solution was the 4\textsuperscript{th} that attained a mean nDCF of \textbf{0.54} with the combination of predictions made by two different pairs of XGBoost classifiers and feature sets for the coarse grain and fine grain data, respectively. Although the overall mean nDCF is lower than the state-of-the-art, the proposed solution outperforms the state-of-art on the coarse grain data with an nDCF of \textbf{0.33} as opposed to the state-of-the-art model's coarse-grain data nDCF of \textbf{0.37}. Table \ref{tab:tab5} shows the NIST post-challenge leaderboard, as of date, with the performances over the different distances where the \textbf{model proposed by this paper}, referred to as \textbf{TsFeX}, is the best performing model that effectually performs contact tracing using \textbf{T}ime \textbf{s}eries \textbf{F}eature \textbf{e}xtraction and the \textbf{X}GBoost model tuned with Bayesian optimization.
\par 
The increase in performance depicts that adding new data by performing an appropriate feature engineering over the set of available features enables the model to achieve better results.

\begin{table}[!ht]
\centering
\resizebox{0.48\textwidth}{!}{%
\begin{tabular}{l|l|l|l|l|l}
Rank &
  Model &
  \begin{tabular}[c]{@{}l@{}}nDCF \\ ($D=1.2\vert$\\ $Set=Fine$)\end{tabular} &
  \begin{tabular}[c]{@{}l@{}}nDCF\\ ($D=1.8\vert$\\ $Set=Fine$)\end{tabular} &
  \begin{tabular}[c]{@{}l@{}}nDCF\\ ($D=3.0\vert$\\ $Set=Fine$)\end{tabular} &
  \begin{tabular}[c]{@{}l@{}}nDCF\\ ($D=1.8\vert$\\ $Set=Coarse$)\end{tabular} \\ \hline
1 & \textbf{TsFeX} & \textbf{0.62} & \textbf{0.60} & \textbf{0.62} & \textbf{0.33} \\ \hline
2 & GBM & 0.60 & 0.52 & 0.58 & 0.37 \\ \hline
3 & LCD & 0.69 & 0.83 & 0.90 & 0.81
\end{tabular}%
}
\vspace{3mm}
\caption{NIST Contact Tracing Post-challenge Leaderboard.}
\label{tab:tab5}
\end{table}

\vspace{-5mm}
\section{Conclusion}

With the aim of containing the spread of COVID-19 pandemic, automated contact tracing has been deemed as a plausible countermeasure. As such, this paper describes the consequential approaches followed in arriving at a gradient boosted classifier model that effectively predicts whether an individual has been in close-contact to someone who has been infected with the virus. The model harnesses inertial measurement unit (IMU) sensor data transmitted from handheld devices to train the model with features extracted from such time series data.

\subsection{Limitations}

The main approaches explored in this work focused on leveraging the temporal information within the data by generating useful statistics using time series feature extraction for tree-based classification. However, such approaches discard temporal information once the features have been created. Furthermore, the XGBoost algorithm's effectiveness at spotting patterns in time dependent data relies heavily upon the number of direct time-related features in the dataset.

\subsection{Future Work}

An alternative approach which could be a promising basis for future work concerning this limitation stated above was considered. In order to more effectively train on time series data, ideally, the model should process the data in a sequential manner. To handle a multivariate dataset comprising of differently sized contact event files, a recurrent neural network model was explored and the recording methodology of the NIST dataset was reconsidered during the pre-processing. As mentioned previously, features in a given file were recorded sporadically within 4 second looks. These intervals were isolated and transformed into a sequence of distinct measurements used to train a sequential classifier. The points where successive measurements occurred at least 10 seconds apart were identified in order to recognize the gaps between looks. The values of each feature within the intervals were then collected and averaged to create a single sequence of inputs corresponding to a sequence of the average measurements. This method of pre-processing better encodes the noticeable change in feature values after a pause in recording and also attempts to capture the dependency of the IMU data on the relative position of the receiving device. 
\par 

The main limitation to this approach is the quality of the dataset itself. The prevalence of shallow sequences shown previously in Figure \ref{fig:fig1} indicates that the model may find it difficult to learn from this data and it is expected that a properly structured dataset with consistent number of intervals would enable more robust learning using this approach. As such, future work in this direction could consider approaches to mitigate the effect of this variance in the number of measurement intervals on the model. Another important aspect to consider is the engineering of position and time dependent features to improve training. In this manner, a model can be trained to infer the movement characteristics of the receiver relative to the transmitter and the distance interval can be estimated using the memory of previous states. In doing so, it may be beneficial to take a more fine grained approach and transform the 4 second measurement intervals into sequences themselves as opposed to taking averages.

\section*{Acknowledgment}
This work was funded by Science Foundation Ireland through the SFI Centre for Research Training in Machine Learning (18/CRT/6183). We extend our sincere appreciation to Sidra Aleem for her constructive mentoring during this project and Dr. Georgiana Ifrim for sharing valuable insight on the time series approach. 

\footnotesize\bibliography{main.bib}

\end{document}